\documentclass[10pt, a4paper]{article}

\usepackage{lrec-coling2024} 

\usepackage{graphicx,xcolor}
\usepackage{url}
\usepackage{array}
\usepackage{here}
\usepackage{float}
\usepackage{multicol}

\usepackage{multirow,tabularx}
\usepackage{array}
\usepackage{here}
\usepackage{float}
\usepackage{multicol}
\usepackage{arydshln}

\usepackage{amsmath}
\usepackage{amssymb}
\usepackage{amsfonts}
\usepackage{latexsym}
\usepackage{comment}
\usepackage{color}

\newcommand{\tabH}{\rule{0pt}{2.1ex}}
\newcommand{\bhline}{\noalign{\hrule height 1.2pt}}
\newcommand{\footlink}[1]{\footnote{\url{#1}}}

\newcommand{\para}[1]{{\vspace{2ex}\noindent\textbf{#1}\hspace{0.5ex}}}
\newcommand{\parapara}[2]{{\vspace{2ex}\noindent\textbf{#1}\hspace{0.75ex}{#2}\hspace{0.75ex}}}
\newcommand{\bgred}[1]{{\colorbox[rgb]{0.99,0.8,0.8}{#1}}}
\newcommand{\bgredbf}[1]{{\colorbox[rgb]{0.99,0.8,0.8}{\textbf{#1}}}}

\title{
WikiSplit++:\\
Easy Data Refinement for Split and Rephrase
}

\name{
Hayato Tsukagoshi$^{\diamondsuit}$, \ \ Tsutomu Hirao$^{\clubsuit}$, \ \ Makoto Morishita$^{\clubsuit}$\\
{\bf \large
Katsuki Chousa$^{\clubsuit}$, 
Ryohei Sasano$^{\diamondsuit}$, \ \ Koichi Takeda$^{\diamondsuit}$
}}

\address{
$^{\diamondsuit}$Graduate School of Informatics, Nagoya University,\ \ \ \ $^{\clubsuit}$NTT Communication Science Laboratories \\
tsukagoshi.hayato.r2@s.mail.nagoya-u.ac.jp,\\
\{tsutomu.hirao,\ makoto.morishita,\ katsuki.chousa\}@ntt.com\\
\{sasano,\ takedasu\}@i.nagoya-u.ac.jp\\
\\}

\abstract{
The task of Split and Rephrase, which splits a complex sentence into multiple simple sentences with the same meaning, improves readability and enhances the performance of downstream tasks in natural language processing (NLP).
However, while Split and Rephrase can be improved using a text-to-text generation approach that applies encoder-decoder models fine-tuned with a large-scale dataset, it still suffers from hallucinations and under-splitting.
To address these issues, this paper presents a simple and strong data refinement approach.
Here, we create WikiSplit++ by removing instances in WikiSplit where complex sentences do not entail at least one of the simpler sentences and reversing the order of reference simple sentences.
Experimental results show that training with WikiSplit++ leads to better performance than training with WikiSplit, even with fewer training instances.
In particular, our approach yields significant gains in the number of splits and the entailment ratio, a  proxy for measuring hallucinations.
\\ \newline  }
\begin{document}

\maketitleabstract

\section{Introduction}

Simplifying complex text without changing its meaning can be achieved through text simplification.
This task involves word deletion, reordering, and insertion, as well as syntactic reconstructions, and it can help reduce the burden of reading for humans or assist with downstream NLP tasks.
Currently, automatic text simplification methods rely on encoder-decoder models~\cite{nisioi-etal-2017-exploring,martin-etal-2020-controllable,devaraj-etal-2022-evaluating}, particularly pre-trained ones like BART~\cite{BART} and T5~\cite{T5}. 

Split and Rephrase, a text simplification task proposed by \citet{SplitAndRephrase}, breaks down a complicated sentence into shorter, simpler ones as much as possible without altering the vocabulary or meaning of the complex sentence.
Most Split and Rephrase methods use encoder-decoder models to accomplish this as a sequence-to-sequence generation task.
Large-scale training datasets, such as WebSplit~\cite{SplitAndRephrase} from the WebNLG corpus~\cite{WebNLG} and WikiSplit~\cite{WikiSplit} from Wikipedia edit histories, are automatically generated and used for the training. 

\begin{table*}[t]
\small
\centering
\begin{tabular}{p{0.36\linewidth}p{0.57\linewidth}}
\bhline
\tabH Complex sentence & Simple sentences \\
\bhline
\multirow{2}{\linewidth}{\tabH Her father was a physician and she was raised in a secular environment.} & \tabH Her father was a physician, and \bgredbf{she followed in his footsteps.}\\
& She was raised in a secular environment. \\
\hline
\multirow{3}{\linewidth}{\tabH It debuted at number 24 on the US ``Billboard'' 200, and at number 70 in Canada.} & \tabH {It debuted at number 24 on the ``Billboard'' 200, \bgredbf{one of} \bgredbf{the top debuts of that week.}}\\
\tabH & The album debuted at number 70 in Canada. \\
\hline
\multirow{2}{\linewidth}{\tabH A pink Hippo-like diplodorian, he can produce bubbles from his mouth.} & \tabH A pink Hippo-like diplodorian.\\
& A \bgredbf{blue diplodorian} who can produce \bgredbf{staples} from his mouth.\\
\bhline
\end{tabular}
\caption{
Actual examples containing hallucinations in WikiSplit~\cite{WikiSplit} are \bgred{marked in red}.
The proposed filtering method using NLI classification removes these unsuitable examples and improves models.
}
\label{tab:hallucination}
\end{table*}

Although current Split and Rephrase methods have been improved, they still have limitations;
they sometimes generate simple sentences with hallucinations and fail to split complex sentences.
Table~\ref{tab:hallucination} presents examples of hallucinations in WikiSplit.
Hallucinations, defined as the generation of unfaithful or nonsensical text~\cite{HallucinationSurvey}, are commonly observed in natural language generation and may be caused by low-quality training datasets, as illustrated in the table.
\footnote{
According to \protect\citet{HallucinationSurvey}, hallucinations can be further classified into two types: intrinsic hallucinations, defined as ``the generated output that contradicts the source content,'' and extrinsic hallucinations, defined as ``the generated output that cannot be verified from the source content.''
The examples in Table~\ref{tab:hallucination} and Figure\ref{fig:nli-filtering} show training instances that can lead to extrinsic hallucination.
}

While automatic data construction methods help create large-scale training datasets, they may also contribute to such errors.
Furthermore, Split and Rephrase methods sometimes fail to split complex sentences, while humans can easily do so.
Encoder-decoder models lack a mechanism to penalize non-splitting explicitly.
When sequences of multiple simple sentences are similar to a complex sentence, encoder-decoder models might produce complex input sentences without modification.
This is because even if the output of the model matches the input, the loss value can still be low during training.

To address these issues, we propose a simple and practical dataset refinement approach.
We start by removing unreliable training examples using a Natural Language Inference (NLI) classifier. Specifically, we exclude pairs of complex sentences and their corresponding shorter ones from the training dataset when the meaning of the complex sentence contradicts that of the shorter ones.
Then, we reverse the order of simple sentences for a complex sentence to serve as a reference for training.
This ensures that the token sequence of a complex sentence becomes more dissimilar to that of its shorter versions.
By applying this approach to WikiSplit, we created a new dataset called \textbf{WikiSplit++}. 
Our experimental results obtained from manually generated benchmark datasets, such as HSplit~\cite{HSplit}, Wiki-BM, and Cont-BM~\cite{SmallButMighty}, demonstrate the effectiveness of our approach.
We also found that T5 fine-tuned with WikiSplit++ successfully suppressed hallucinations in simple sentences and produced more splits than that trained with WikiSplit.

Our contributions are as follows\footnote{
Our code and dataset are publicly available at \url{https://github.com/nttcslab-nlp/wikisplit-pp} and \url{https://huggingface.co/datasets/cl-nagoya/wikisplit-pp}.
}:
\begin{enumerate}
\item 
We propose a simple and practical data refinement approach using NLI classification and reversing the order of simple sentences for the Split and Rephrase task.
We also construct WikiSplit++, a valuable resource for researchers in the Split and Rephrase research community.
\item 
We demonstrate that T5 trained with WikiSplit++ produces fewer hallucinations and more splits than that trained with WikiSplit.
\end{enumerate}

\section{Related Work}

\subsection{Methodologies}
The Split and Rephrase technique aims to break down complex sentences into simpler and shorter ones as much as possible without changing the vocabulary or meaning. 
It is this point that makes Split and Rephrase differ from other text simplification tasks.
Since we can treat Split and Rephrase as a sequence-to-sequence generation task, we can usually use encoder-decoder models.
\citet{SplitAndRephrase} introduced a multiple source sequence-to-sequence approach using a three-layer LSTM for the task.
It receives a complex sentence and a Resource Description Format (RDF) tuple of the origin and generates simple sentences. 
\citet{BetterEvaluation} introduced a copy mechanism~\cite{CopyMechanism} to improve the model.
Recently, \citet{BiSECT} introduced adaptive loss using distant supervision to further enhance the model and achieve state-of-the-art performance.

\subsection{Datasets}
In order to train Split and Rephrase models, researchers have developed large-scale datasets.
The first benchmark dataset for Split and Rephrase, called WebSplit~\cite{SplitAndRephrase}, was collected from the WebNLG~\cite{WebNLG} dataset, which includes RDF tuples and their corresponding texts.
However, there are many overlaps in the original WebSplit dataset among the train, dev, and test datasets, so \citet{BetterEvaluation} removed them and showed that this cleaning process improved Split and Rephrase.
Despite this improvement, WebSplit remains limited to containing unnatural linguistic expressions with a small vocabulary.
To address this, \citet{WikiSplit} constructed WikiSplit, which was automatically obtained from Wikipedia edit history.
Recently, \citet{MinWikiSplit} constructed MinWikiSplit, a dataset that applies DisSim~\cite{DisSim}, a rule-based sentence splitting system, to WikiSplit.
\citet{BiSECT} proposed BiSECT, which consists of one-to-two sentence pairs extracted from parallel corpora and translated into the same language to construct a Split and Rephrase dataset.
These datasets are large enough to train encoder-decoder models, but, as mentioned earlier, they include inappropriate pairs of complex sentences and their corresponding simple sentences.

On the other hand, there are small yet valuable datasets that are manually created for evaluation.
One example is HSplit~\cite{HSplit}, comprised of complex sentences from Wikipedia and their corresponding simple sentences.
\citet{SmallButMighty} also created Wiki-BM and Cont-BM, which offer a wider range of vocabulary and syntactic structures based on sentences from Wikipedia and contracts. 
Although these datasets may be insufficient for the training of encoder-decoder models, they provide valuable resources for the evaluation of Split and Rephrase systems.

\subsection{Data Refinement}

To improve the quality of natural language generation, it is crucial to refine the training datasets.
While naive data refinement, including elimination of duplications, has been proposed, linguistically motivated methods have also been introduced.
Specifically, textual entailment has been used as a key indicator.

\citet{carpuat-detecting} came up with a data refinement method for parallel corpora that excludes inappropriate translation pairs.
Their approach measures the quality of translations based on cross-lingual textual entailment and additional length-based features.
Accordingly, their method eliminates unreliable translation pairs based on this quality assessment. 
Similarly, \citet{ImprovingTruthfulness} used textual entailment to eliminate pairs of untruthful headlines and corresponding source articles.
They fine-tuned an existing NLI classification model with their own training data since source articles have multiple sentences.
Meanwhile, \citet{Deduplicating} demonstrated that excluding duplication in the training data is effective for training language models.
These methods mainly focus on getting rid of unsuitable pairs of source and target sentences, that is, they do not consider modifying the training data to enhance generation quality.

\section{Proposed Method}

First, we eliminate any pair of complex and simple sentences that have conflicting meanings.
Second, we reverse the order of the simple sentences and serve them as reference supervision in the training dataset.

\subsection{Natural Language Inference Classification}

\begin{figure}
    \centering
    \centering
    \includegraphics[keepaspectratio, width=\linewidth]{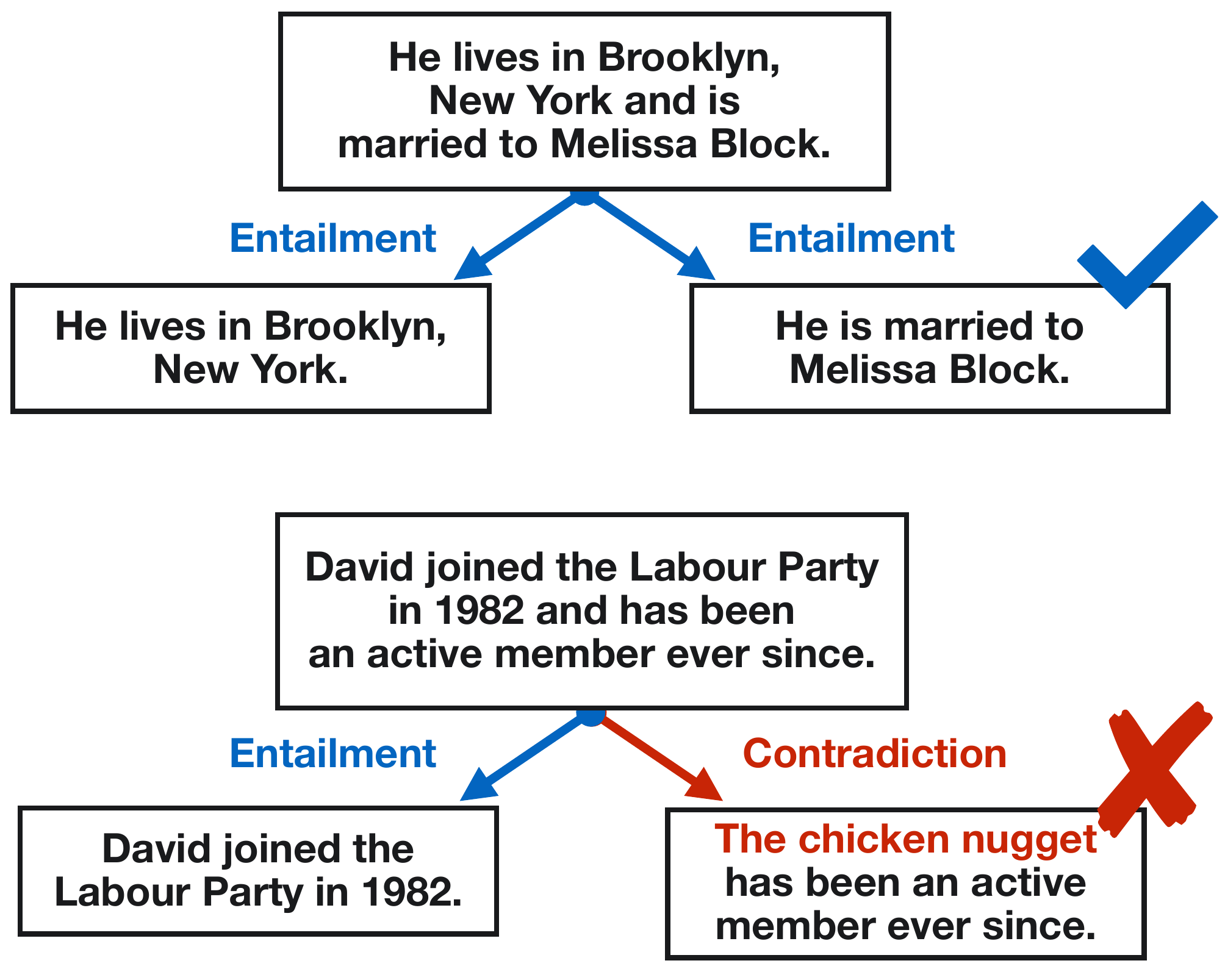}
    \caption{Overview of NLI classification. These are actual examples from WikiSplit.}
    \label{fig:nli-filtering}
\end{figure}

To suppress the production of simple sentences that contradict complex ones, it is important to remove inappropriate sentence pairs (like those in Table~\ref{tab:hallucination}) from the dataset used to train a Split and Rephrase model.
Natural Language Inference (NLI) classification models can be useful for identifying these contradictions.
The NLI classification models classify the relationship between a given premise and a hypothesis sentence into entailment, contradiction, or neutral.
The Stanford NLI (SNLI) dataset~\cite{SNLI} and the Multi-Genre NLI (MNLI) dataset~\cite{MultiGenreNLI} are commonly used to train NLI classification methods.

We show a procedure to remove inappropriate pairs of sentences in Figure~\ref{fig:nli-filtering}. 
When a complex sentence $c$ and the corresponding simple sentences $s_1,...,s_n$ are given, we put each pair of complex sentence $c$ and simple sentence $s_i\ (1 \leq i \leq n)$ into the NLI classification model.
Here, $P_\mathrm{ent}$ denotes the probability that a pair of sentences belongs to \textit{entailment}.
We regard pairs of sentences as appropriate when probability $P_\mathrm{ent}$ is higher than that of contradiction or neutral.
When all simple sentences pass the above condition, we employ pairs of complex sentences and corresponding simple ones in the training dataset.

Unlike the settings of \citet{ImprovingTruthfulness}, which identify contradictions between an article and its headline, we focus on identifying contradictions between each sentence pair.
Therefore, we simply employ an existing NLI model without any modification.

\subsection{Sentence-Order Reversing}

In many cases, the complex sentence and corresponding simple sentences result in similar sequences of tokens.
During training, to generate the sentence, the encoder-decoder model is trained with cross-entropy loss to assess the tokens at a particular time $t$.
In order to split a sentence, a period must be output in the appropriate position.
However, there are cases where it fails to do so, i.e., outputting an input sentence as is.
Because the period as a symbol for splitting a sentence is not given special treatment, the loss value is significantly low in such cases.
Additionally, the lower frequency of periods compared to other words in the input sentence is another contributing factor for the decoder to reproduce the input complex sentences.

\begin{figure}
    \centering
    \includegraphics[keepaspectratio, width=\linewidth]{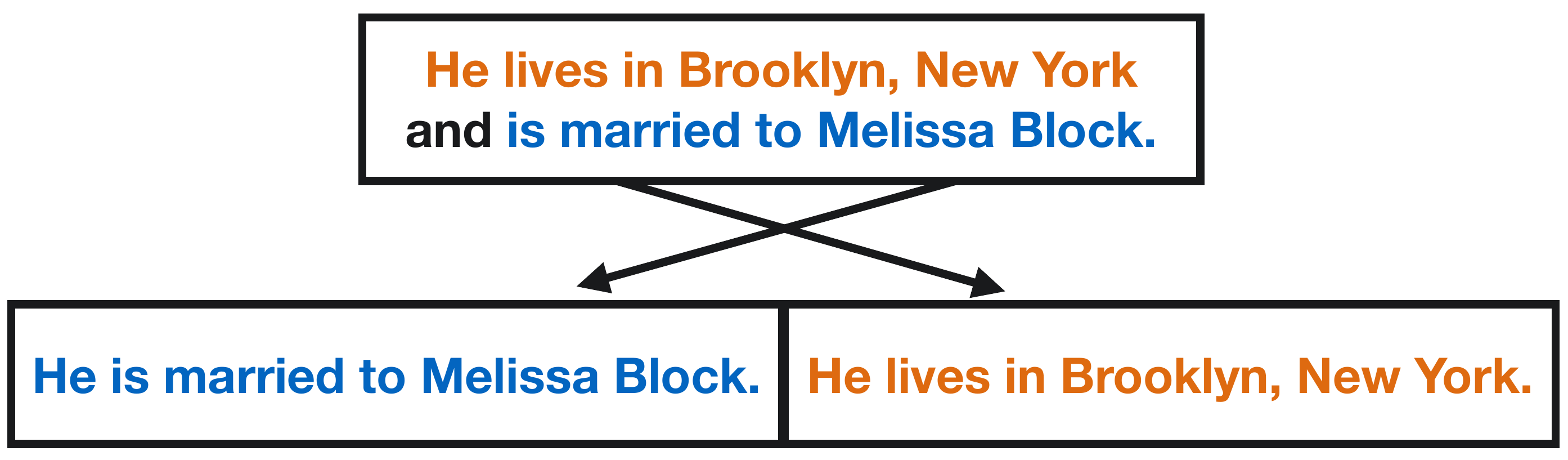}
    \caption{
Overview of sentence-order reversing.
To prevent the model from outputting the sentences as they are, the model is trained with the reverse order of the sentences as the gold labels.
}
    \vspace{2ex}
    \label{fig:sentence-order-reversing}
\end{figure}

To address this issue and ensure that the models successfully split complex sentences, we can simply disrupt the similarity between the input and output sequences.
One simple way to do this is by reversing the order of simple sentences as shown in Figure~\ref{fig:sentence-order-reversing}.
By doing this, the models are prevented from simply reproducing the input sentences word-for-word.

It is possible for the first sentence in a split to contain named entities (NEs) that are later referred to using pronouns.
Reversing the order of split sentences may seem to compromise the relationship between NEs and pronouns.
However, because Transformer-based encoders can encode sequences bidirectionally, the reverse order does not cause inconsistencies during training.

\section{Experimental Settings}

\subsection{Datasets}

We developed WikiSplit++ by applying our data refinement to WikiSplit~\cite{WikiSplit} because it is the largest dataset and its sentences are written by humans; that is, it is assumed to have more natural examples.
While the size of BiSECT~\cite{BiSECT} is nearly the same as WikiSplit, sentences are automatically generated using machine translation. 
We follow the official train/dev dataset split for WikiSplit, and our data refinement approach is applied for each dataset.
We then use it to train the Split and Rephrase models.

Table~\ref{tab:dataset} shows the statistics of WikiSplit and WikiSplit++.
By applying our data refinement, the number of instances for each dataset
is significantly reduced. 
Approximately 35\% of the sentences were removed through this process.
Note that we do not use the test dataset because we conduct evaluations using other datasets, as shown below.

\begin{table}[t]
\centering
\begin{tabular}{lrr}
\bhline
\tabH & WikiSplit & WikiSplit++\\
\bhline
\tabH Overall & 994,481 & 630,433 \\
train & 795,585 & 504,375 \\
dev & 99,448 & 63,065 \\
test & 99,448 & 62,993 \\
\bhline
\end{tabular}
\caption{Number of instances in the datasets used for the experiment}
\label{tab:dataset}
\end{table}

\parapara{HSplit}{\cite{HSplit}} contains 359 complex sentences obtained from Wikipedia and each complex sentence has simple sentences generated by four different annotators.

\parapara{Wiki-BM}{\cite{SmallButMighty}} contains 500 complex sentences obtained from the test dataset of WikiSplit. 
To create references for simple sentences, three annotators generated them for each complex sentence.
Then, two other annotators judged which simple sentences were `perfect.'
We used these as references.

\parapara{Cont-BM}{\cite{SmallButMighty}} contains 500 complex sentences obtained from publicly available legal procurement contracts. Reference simple sentences were created in the same manner as Wiki-BM.

\subsection{Evaluation Metrics}

\paragraph{Automatic Evaluation}
We utilized BLEU~\cite{BLEU}, BERTScore~\cite{BERTScore}, SARI~\cite{SARI}, and Flesch-Kincaid Grade Level (FKGL) for automatic metrics by following previous studies.
We also included BLEURT~\cite{BLEURT}, which has been reported to have a better correlation with human ratings than BERTScore.
BLEU, BERTScore, BLEURT, and SARI measure the quality of the generated sentences in terms of their content.
On the other hand, FKGL measures the readability of generated sentences, where smaller scores indicate good readability.
We also compute the average number of splits and the percentage of outputs identical to the complex.

Furthermore, we compute the ratio of complex sentences that entail corresponding simple sentences based on NLI classification as a proxy for the evaluation of hallucinations.
We refer to this metric as the \textbf{`Entailment ratio'}.
It has been shown that such NLI scores, as a measure of factual consistency, have a higher correlation with human evaluation than metrics that measure word overlap~\cite{Q2}, suggesting that they can also accurately assess the presence of hallucinations~\cite{kryscinski-etal-2020-evaluating}.
Note that the NLI classification model is the same as that used for data refinement.

\subsection{Baseline Methods}

To investigate the impact of WikiSplit++, we compare our methods with the following baselines.

\para{Echo} outputs the input complex sentence as is.

\parapara{DisSim}{\cite{DisSim}} is a discourse-aware sentence-splitting framework based on hand-crafted rules that splits a complex sentence recursively by applying a small set of 35 rules.

\parapara{BiSECT Model}{\cite{BiSECT}\protect\footnote{
We refer to the Split and Rephrase method proposed by \citet{BiSECT} as the BiSECT Model and the dataset described in the paper as BiSECT.
}} is a SOTA Split and Rephrase method based on BERT-Initialized Transformer~\cite{BERT-initialized}.
We used a publicly available pre-trained model\footlink{https://github.com/mounicam/BiSECT} trained with BiSECT and WikiSplit datasets.

\para{GPT-3} is a pre-trained language model with an enormous number of parameters, and it has achieved top performance in natural language generation tasks.
We conducted experiments using both zero-shot and three-shot learning settings and utilized the \texttt{text-davinci-003} model from OpenAI.
The prompts we used are as follows:
\begin{quote}
    \textit{Split and rephrase the following complex sentence into a simple and concise number of sentences, while maintaining the structure, phrases, and meaning of the sentences.}\\
    \textbf{\texttt{Complex sentence}}: \textit{An example of the complex sentence}\\
    \textbf{\texttt{Simple sentences}}: \textit{The reference simple sentences.}\\
\end{quote}

\begin{table*}[t]
\centering
\small
\begin{tabular}{cc@{\hspace{6ex}}ccccc}
\bhline
\multirow{2}{*}{NLI} & \multirow{2}{*}{Rev.} & \multicolumn{5}{c}{\tabH Learning Rate} \\
 & & 5e-5 & 1e-4 & 2e-4 & 5e-4 & 1e-3 \\
\bhline
\tabH & & 0.6453 & 0.6356 & 0.6267 & 0.6188 & \textbf{0.6167} \\
\checkmark & & 0.3319 & 0.3245 & 0.3192 & \textbf{0.3146} & 0.3147\\
& \checkmark & 0.6523 & 0.6416 & 0.6324 & 0.6238 & \textbf{0.6211} \\
\checkmark & \checkmark & 0.3377 & 0.3295 & 0.3233 & 0.3187 & \textbf{0.3182} \\
\bhline
\end{tabular}
\caption{
Validation losses for each learning rate.
``NLI'' indicates that filtering by NLI classification has been applied, and ``Rev.'' indicates that sentence-order reversing has been applied.
The loss values are the averages of three experiments with different random seed values.
}
\vspace{1ex}
\label{tab:tuning-hyperparameters}
\end{table*}

\subsection{Implementation Details}

\paragraph{Encoder-Decoder Model}
We used T5-small~\cite{T5}, a popular pre-trained encoder-decoder model for sequence-to-sequence generation.
We implemented our experiments with PyTorch~\cite{PyTorch} and downloaded the pre-trained checkpoint of T5-small from HuggingFace's Transformers~\cite{Transformers}.

\paragraph{NLI Classification}
We used DeBERTa-v2 XXL~\cite{DeBERTa} fine-tuned on MNLI\footlink{https://huggingface.co/microsoft/deberta-v2-xxlarge-mnli} as the pre-trained NLI classification model.
After applying our proposed filtering to WikiSplit, the total number of cases went from 994,481 to 630,433, thus removing 364,048 cases, or about 36.6\% of the entire dataset.
The dataset was then divided into train/dev/test sets at a ratio of 8:1:1.
The number of cases in the dataset before and after 
filtering with NLI classification is shown in Table~\ref{tab:dataset}.

\paragraph{Sentence-Order Reversing}
To reverse the order of simple sentences, we have to identify the sentence boundaries in a sequence of multiple simple sentences.
The period is a significant cue for identifying the sentence boundaries; however, we failed to split sentences when using the period as the only cue.
Therefore, we applied PySBD~\cite{PySBD} to the raw string in the datasets to split them.
PySBD is a rule-based sentence boundary detector that outperforms conventional sentence boundary detection tools such as NLTK and SpaCy.
The boundary detection accuracy is about 97\% on the GENIA corpus~\cite{read-etal-2012-sentence}.

Please note that the models generate shorter (simplified) sentences in reverse order as one sequence of tokens.
To identify these shorter sentences from the sequence, we detected sentence boundaries within this sequence by utilizing PySBD.
Once identified, we rearranged the sentences in reverse order to obtain the final result. 

\paragraph{Fine-tuning}

We set the batch size to 32 and the number of training steps to 20,000, and we used the AdamW~\cite{AdamW} optimizer with a linear learning rate warmup of 2,000 steps.
The learning rate was selected from \{1e-3, 5e-4, 2e-4, 1e-4, 5e-5\} using the development set.
We show the losses obtained from the development dataset in Table~\ref{tab:tuning-hyperparameters}.
We ran the experiment three times with different random seed values for each learning rate and selected the one with the smallest average loss.
During inference, we used a 10-width beam-search and ensured that the beam-search did not repeat trigrams.
We evaluated the models on the development set every 1,000 steps and used the checkpoint with the smallest loss on the development set for evaluation.

For fine-tuning, we used a single NVIDIA A6000 GPU and BFloat16 data type.
The training time for the T5-small model was about 70 minutes for the WikiSplit dataset and about 60 minutes for the WikiSplit++ dataset.
Consequently, dataset filtering facilitated a training acceleration of approximately 15\%\footnote{The degree of speedup is smaller than the reduction in the number of training examples (36.6\%) due to various overheads, including evaluations on the development set.}.

\paragraph{Evaluation}
We used the pre-trained model published by \citet{BiSECT} on GitHub\footlink{https://github.com/mounicam/BiSECT} for comparison.
Since the output simple sentences from DisSim and BiSECT are tokenized into words, a detokenize process was applied for a fair comparison.
The detokenize process was applied to our model and the same process was applied to the training and evaluation datasets.
We used sacreBLEU~\cite{SacreBLEU} to compute corpus-level and sentence-level BLEU.
A model trained by applying sentence-order reversing produces reversed simple sentences, which is inconvenient for evaluation.
Therefore, we also reversed the output order of simple sentences from these models during inference.

\begin{table*}[ht]
\centering
\tabcolsep 3pt
\small
\begin{tabular}{llcccccccc}
\bhline
\tabH System & Dataset & \ BLEU\ & BERTScore & BLEURT  & SARI & Entailment & FKGL & \#Sent. & \ Copy\ \\
\bhline
\multicolumn{10}{c}{\tabH HSplit~\cite{HSplit}} \\
\hline
\tabH Echo & N/A & \ \ 88.91 & \ \ 97.10 & 84.48 & 66.60 & 100.00 & 12.81 & 1.00 & 100.00 \\
DisSim & N/A & \ \ 63.71 & \ \ 94.93 & 75.54 & 66.74 & 92.20 & 7.85 & 2.96 & \ \ 21.73 \\
BiSECT Model & BiSECT+WikiSplit & \ \ 86.98 & \ \ 96.54 & 81.35 & 57.65 & 95.26 & 8.58 & 1.98 & \ \ \ \ 2.23 \\
GPT-3 (zero-shot) & N/A & \ \ 54.39 & \ \ 94.25 & 76.79 & 74.34 & 94.15 & 8.90 & 2.14 &\ \ \ \  5.29 \\
GPT-3 (3-shot) & N/A & \ \ 72.09 & \ \ 95.79 & 79.66 & 69.46 & 96.66 & 9.09 & 1.86 & \ \ 18.94 \\
\hline
\tabH T5-small & WikiSplit & \ \ 87.95 & \ \ 96.65 & 82.06 & 57.17 & 95.49 & 8.63 & 1.98 & \ \ \ \  2.48 \\
T5-small & WikiSplit++ & \ \ 88.06 & \ \ 96.57 & 81.71 & 56.79 & 98.02 & 8.59 & 2.00 &\ \ \ \  0.72 \\
\hline
\tabH Reference & N/A & 100.00 & 100.00 & 89.81 & 57.26 & 99.16 & 8.83 & 1.98 &\ \ \ \  N/A \\
\bhline
\multicolumn{10}{c}{\tabH Wiki-BM~\cite{SmallButMighty}} \\
\hline
\tabH Echo & N/A & \ \ 71.72 & \ \ 97.91 & 83.07 & 62.78 & 100.00 & 14.81 & 1.01 & 100.00 \\
DisSim & N/A & \ \ 56.86 &\ \  95.76 & 74.08 & 57.66 & 86.60 & 7.33 & 4.18 & \ \ \ \  6.20 \\
BiSECT Model & BiSECT+WikiSplit & \ \ 78.05 & \ \ 98.31 & 85.18 & 43.37 & 98.26 & 9.59 & 2.00 &\ \ \ \  0.74 \\
GPT-3 (zero-shot) & N/A & \ \ 58.71 & \ \ 96.42 & 80.81 & 66.45 & 96.77 & 8.88 & 2.40 & \ \ \ \  0.50 \\
GPT-3 (3-shot) & N/A & \ \ 71.14 &\ \  97.89 & 84.05 & 57.05 & 98.51 & 9.15 & 2.17 &\ \ \ \  1.49 \\
\hline
\tabH T5-small & WikiSplit & \ \ 78.26 & \ \ 98.34 & 85.42 & 42.96 & 99.01 & 9.57 & 2.00 & \ \ \ \ 0.67 \\
T5-small & WikiSplit++ & \ \ 78.12 & \ \ 98.32 & 85.28 & 43.37 & 99.13 & 9.55 & 2.00 &\ \  \ \ 0.97 \\
\hline
\tabH Reference & N/A & 100.00 & 100.00 & 93.39 & 37.39 & 98.76 & 7.92 & 3.06 & \ \ \ \ N/A \\
\bhline
\multicolumn{10}{c}{\tabH Cont-BM~\cite{SmallButMighty}} \\
\hline
\tabH Echo & N/A & \ \ 72.65 & \ \ 97.19 & 83.42 & 63.06 & 99.75 & 21.87 & 1.03 & 100.00 \\
DisSim & N/A & \ \ 54.67 & \ \ 94.43 & 71.23 & 61.69 & 68.47 & 10.53 & 4.37 & \ \ \ \ 8.62 \\
BiSECT Model & BiSECT+WikiSplit & \ \ 68.52 & \ \ 96.54 & 78.87 & 57.72 & 90.15 & 14.17 & 1.99 & \ \ \ \ 4.68 \\
GPT-3 (zero-shot) & N/A & \ \ 49.96 & \ \ 94.31 & 76.88 & 75.31 & 93.84 & 13.49 & 2.38 & \ \ \ \ 2.96 \\
GPT-3 (3-shot) & N/A & \ \ 58.76 &\ \  95.89 & 79.37 & 70.69 & 94.09 & 13.94 & 2.10 & \ \ 12.07 \\
\hline
\tabH T5-small & WikiSplit & \ \ 72.31 & \ \ 97.18 & 81.89 & 51.40 & 96.92 & 14.32 & 1.94 & \ \ \ \ 8.42 \\
T5-small & WikiSplit++ & \ \ 71.74 & \ \  97.06 & 81.26 & 51.72 & 96.35 & 14.30 & 1.99 & \ \ \ \ 3.47 \\
\hline
\tabH Reference & N/A & 100.00 & 100.00 & 93.59 & 37.95 & 92.61 & 11.67 & 3.13 & \ \ \ \ N/A \\
\bhline
\end{tabular}
\caption{
Automatic evaluation results on HSplit, Wiki-BM, and Cont-BM.
Values are reported for corpus-level BLEU (BLEU), BERTScore, BLEURT, SARI, Entailment ratio (Entailment), FKGL, the average number of sentences (\#Sent.), and the percentage of output equal to input (Copy).
All scores except ``\#Sent.'' are multiplied by 100.
}
\label{tab:automatic-experiment}
\end{table*}

\begin{table*}[ht]
\small
\renewcommand{\arraystretch}{1.3}
\centering
\begin{tabular}{p{0.25\linewidth}p{0.69\linewidth}}
\bhline
\tabH \multirow{2}{*}{Input} &  In such event, IBM reserves the right to modify the terms of the Special Bid or to cancel your Special Bid authorisation.\\
\hline
\tabH \multirow{2}{*}{Reference} &  In such an event, IBM reserves the right to modify the terms of the Special Bid. IBM can also cancel your Special Bid authorisation.\\
\bhline
\tabH System & Output \\
\bhline
\tabH \multirow{2}{*}{DisSim} & IBM reserves the right to modify the terms of the Special Bid or to cancel your Special Bid authorisation. This is in such event. \\
\hline
\multirow{2}{*}{BiSECT Model} & In such event, IBM reserves the right to modify the terms of the Special Bid. \bgred{You can also cancel} your special bid authorisation. \\
\hline
\multirow{2}{*}{T5-small} & In such event, IBM reserves the right to modify the terms of the Special Bid. It \bgred{may also be able to cancel} your special bid authorisation.\\
\hline
\multirow{2}{*}{T5-small + NLI + Rev.} & In such event, IBM reserves the right to modify the terms of the Special Bid. Or to cancel your special bid authorisation. \\
\bhline
\end{tabular}
\caption{
Actual example from Cont-BM~\cite{SmallButMighty} and corresponding outputs from each system.
``NLI'' denotes the model trained on the dataset where NLI classification was applied, and ``Rev.'' denotes the model trained on the dataset where the order of simple sentences was reversed.
}
\vspace{1ex}
\label{tab:generated}
\renewcommand{\arraystretch}{1.0}
\end{table*}

\section{Experimental Results and Discussion}

\subsection{Results of Automatic Evaluation}

\begin{table*}[ht]
\centering
\small
\begin{tabular}{@{\hspace{2ex}}l@{\hspace{6ex}}cccccccc}
\bhline
\tabH Dataset & BLEU & BERTScore & BLEURT  & SARI & Entailment & FKGL & \#Sent. & Copy  \\
\bhline
\multicolumn{9}{c}{\tabH HSplit~\cite{HSplit}} \\
\hline
\tabH WikiSplit & 87.95 & 96.65 & 82.06 & 57.17 & 95.49 & 8.63 & 1.98 & \ \ 2.48 \\
WikiSplit++ & 88.06 & 96.57 & 81.71 & 56.79 & 98.02 & 8.59 & 2.00 & \ \ 0.72 \\
MinWikiSplit & 77.98 & 95.77 & 76.38 & 65.45 & 83.54 & 8.47 & 2.11 & 25.88 \\
MinWikiSplit++ & 77.67 & 95.71 & 76.87 & 65.93 & 90.11 & 8.49 & 2.13 & 27.24 \\
BiSECT & 73.57 & 96.10 & 79.13 & 67.41 & 90.72 & 8.73 & 1.98 & \ \ 2.79 \\
BiSECT++ & 73.54 & 96.01 & 79.44 & 68.13 & 95.88 & 8.57 & 1.99 & \ \ 1.20 \\
\bhline
\multicolumn{8}{c}{\tabH Wiki-BM~\cite{SmallButMighty}} \\
\hline
\tabH WikiSplit & 78.26 & 98.34 & 85.42 & 42.96 & 99.01 & 9.57 & 2.00 & \ \ 0.67 \\
WikiSplit++ & 78.12 & 98.32 & 85.28 & 43.37 & 99.13 & 9.55 & 2.00 & \ \ 0.97 \\
MinWikiSplit & 71.68 & 96.94 & 76.62 & 56.54 & 86.43 & 8.53 & 2.58 & 11.02 \\
MinWikiSplit++ & 70.28 & 96.61 & 75.37 & 57.20 & 90.15 & 8.19 & 2.84 & \ \ 6.80 \\
BiSECT & 67.27 & 97.86 & 82.97 & 60.51 & 98.88 & 9.51 & 1.99 & \ \ 1.36 \\
BiSECT++ & 66.66 & 97.82 & 82.72 & 60.88 & 98.36 & 9.55 & 2.00 & \ \ 1.04 \\
\bhline
\multicolumn{8}{c}{\tabH Cont-BM~\cite{SmallButMighty}} \\
\hline
\tabH WikiSplit & 72.31 & 97.18 & 81.89 & 51.40 & 96.92 & 14.32 & 1.94 & \ \ 8.42 \\
WikiSplit++ & 71.74 & 97.06 & 81.26 & 51.72 & 96.35 & 14.30 & 1.99 & \ \ 3.47 \\
MinWikiSplit & 67.29 & 95.58 & 71.96 & 62.19 & 71.45 & 11.95 & 2.78 & \ \ 6.67 \\
MinWikiSplit++ & 65.70 & 95.36 & 71.17 & 63.54 & 80.86 & 12.01 & 2.76 & \ \ 8.33 \\
BiSECT & 60.65 & 96.05 & 78.11 & 69.39 & 87.29 & 14.47 & 1.96 & \ \ 7.36 \\
BiSECT++ & 59.38 & 95.99 & 77.76 & 70.12 & 94.58 & 14.21 & 2.00 & \ \ 2.98 \\
\bhline
\end{tabular}
\caption{
Ablation studies on other datasets.
Each column has the same meaning as in Table~\ref{tab:automatic-experiment}.
}
\label{tab:ablation-datasets}
\end{table*}

\begin{table*}[t]
\centering
\small
\begin{tabular}{cccccccccc}
\bhline
\tabH NLI & Rev. & BLUE & BERTScore & BLEURT & SARI & Entailment & FKGL & \#Sent. & Copy \\
\bhline
\multicolumn{10}{c}{\tabH HSplit~\cite{HSplit}} \\
\hline
\tabH & & 87.95 & 96.65 & 82.06 & 57.17 & 95.49 & 8.63 & 1.98 & 2.48\\
\checkmark & & 88.81 & 96.80 & 82.79 & 57.23 & 97.74 & 8.71 & 1.93 & 7.35\\
& \checkmark &  88.21 & 96.58 & 81.68 & 56.68 & 96.74 & 8.57 & 2.00 & 0.33\\
\checkmark & \checkmark & 88.06 & 96.57 & 81.71 & 56.79 & 98.02 & 8.59 & 2.00 & 0.72\\
\bhline
\end{tabular}
\caption{
Results of evaluating how each technique affects performance
}
\vspace{1ex}
\label{tab:ablation-technique}
\end{table*}

\begin{table*}[ht]
\centering
\small
\begin{tabular}{lcccccccc}
\bhline
\tabH Classifier & BLUE & BERTScore & BLEURT & SARI & Entailment & FKGL & \#Sent. & Copy \\
\bhline
\multicolumn{9}{c}{\tabH HSplit~\cite{HSplit}} \\
\hline
\tabH N/A  & 87.95 & 96.65 & 82.06 & 57.17 & 95.49 & 8.63 & 1.98 & 2.48 \\
DeBERTa & 88.06 & 96.57 & 81.71 & 56.79 & 98.02 & 8.59 & 2.00 & 0.72 \\
RoBERTa & 88.08 & 96.57 & 81.74 & 56.73 & 98.25 & 8.60 & 2.00 & 0.67 \\
TRUE & 88.20 & 96.58 & 81.79 & 56.73 & 98.25 & 8.61 & 2.00 & 0.45 \\
\bhline
\end{tabular}
\caption{Results of using different NLI classifiers for filtering datasets}
\label{tab:classifier-experiment}
\end{table*}

Table~\ref{tab:automatic-experiment} shows the automatic evaluation results. 
We found that the performances differed among the datasets.
Scores on Cont-BM are lower than those on HSplit and Wiki-BM.
In particular, the differences among BLEU, SARI, and FKGL are remarkable. 
We believe that the difference in the domains caused these results.
Cont-BM is made from contract documents prepared in their own writing styles, while HSplit and Wiki-BM are made from Wikipedia in a standard writing style.

When compared to the baseline methods, vanilla T5 trained with WikiSplit exhibited superior performance over DisSim across all datasets.
Furthermore, in many instances, it achieved better results than the BiSECT model.
These findings indicate that the current state-of-the-art encoder-decoder model has sufficient capability to handle the Split and Rephrase tasks.

DisSim tended to generate too many simple sentences, which could lead to inaccuracies.
Surprisingly, despite this issue, it achieved the highest scores for SARI and FKGL.
This suggests that these metrics may not be the most reliable way to evaluate the effectiveness of Split and Rephrase.

Based on our evaluation, neither the GTP-3 zero-shot nor the 3-shot model performed well.
In comparison to DisSim, the zero-shot model performed similarly.
Meanwhile, the 3-shot model showed some improvement, but it still fell short when compared to our methods and the BiSECT model. 
Additionally, we found that GTP-3 models tended to have a high number of copy operations but only a moderate number of splits on average.
This indicates that their number of splits can be unstable, leading to both over-split and under-split sentences.

Our proposed methods outperformed the baseline methods in Entailment ratio.
The Entailment ratio scores show significant improvement with the introduction of WikiSplit++.
These results demonstrate the effectiveness of the NLI classification component. 
However, it should be noted that NLI classification may not be as effective when dealing with contract documents. 

Our data refinement approach allowed for the removal of unreliable instances, which led to a smaller training dataset; as mentioned above, 36.6\% of the instances were eliminated.  
Generally, training with a smaller number of instances would not yield significant improvement.
However, our T5 with WikiSplit++ obtained better results than methods with more extensive training data.
As another advantage, the small size of the dataset allowed us to conduct efficient training.

When focusing on the number of splits, it becomes clear that we can generate a greater quantity of simple sentences by introducing WikiSplit++.
We also found that WikiSplit++ successfully suppressed T5's output of sentences identical to the input sentences.
However, there is still room for improvement, since we require further splits in order to achieve the level of human splits.

We present sentence examples in Table~\ref{tab:generated}.
DisSim produced incoherent sentences, while the BiSECT model and T5-small generated fluent sentences but suffered from hallucinations.
In contrast, our model produced relatively high-quality splits.

\subsection{Impact of Data Refinement on other Datasets}

In order to assess the generality of our approach, we conducted experimental evaluations by applying our method to various datasets.
Table~\ref{tab:ablation-datasets} shows the results obtained from MinWikiSplit++ and BiSECT++, which are applied to our proposed methods against MinWikiSplit and BiSECT, respectively.

In similarity-based evaluation metrics such as BLEU, SARI, and so on, data refinement has marginal impact.
The differences in scores are also minor before and after applying the data refinement.
However, we found remarkable gains in the Entailment ratio for all datasets.
The results suggest that our data refinement contributes to reducing generations of simple sentences that are not entailed by the source complex sentences, which cannot be identified by similarity-based metrics.
Furthermore, data refinement improves the number of splits.
These results imply that our data refinement method has a significant impact on the desirable aspects of Split and Rephrase, reducing hallucinations and increasing the number of splits, even though this does not contribute to greatly improving the scores of similarity-based evaluation metrics.

\subsection{Ablation Study of Each Technique}

We employed two techniques for dataset refinement: NLI filtering and sentence-order reversing.
For greater understanding, we conducted an ablation study to assess the impact of each technique.

Table~\ref{tab:ablation-technique} shows the results of experiments conducted with Hsplit.
According to the results, we found that the Entailment ratio consistently improved with the application of NLI filtering.
Although introducing NLI filtering alone reduced the number of splits, the introduction of sentence-order reversing leads to an increase in the number of splits, mitigating the undesired effect of reduced split counts caused by NLI filtering.
Additionally, sentence-order reversing notably decreased the proportion of cases where the input sentence is output as is (i.e., ``Copy'' is reduced).
While sentence-order reversing alone enhanced the Entailment ratio, the employment of NLI filtering yielded a more substantial improvement.
Moreover, the combination of NLI filtering and sentence-order reversing synergistically augmented the Entailment ratio.

\subsection{Impact of NLI Classifier on Performance}

Since our experiments employ the same NLI classifier for both data refinement and system evaluation, it might seem to be an unfair evaluation.
To justify our settings, we provide results of HSplit from using a different NLI classifier for pruning the dataset.
For comparison, we employed TRUE-xxl~\cite{TRUE}---a T5-xxl model fine-tuned on several NLI datasets---and RoBERTa-large~\cite{RoBERTa} fine-tuned on MNLI.
We conducted experiments using three NLI classifiers (DeBERTa, RoBERTa, and TRUE).
For consistency with Table~\ref{tab:automatic-experiment}, we also performed sentence-order reversing.

Table~\ref{tab:classifier-experiment} shows the results.
The results demonstrate that there were consistent improvements in the score for entailment rate, regardless of which NLI classifier was used to prune the dataset. This indicates that our method's enhancements are certainly not due to the use of the same model for both refining and evaluating data.

\subsection{Bias from NLI Classifier}

\begin{table}[t]
\centering
\small
\begin{tabular}{lccc}
\bhline
\tabH  & Entailment & Neutral & Contradiction \\
\bhline
\tabH HSplit & 100.00 & 0.00 & 0.00 \\
Wiki-BM & 100.00 & 0.00 & 0.00 \\
Cont-BM & \ \ 98.77 & 0.49 & 0.74 \\
\bhline
\end{tabular}
\caption{
Proportion of instances classified into each of three labels
}
\label{tab:nli-bias}
\end{table}

We used an NLI classifier for dataset filtering, which raised concern that the classifier's accuracy could introduce bias into the filtered dataset.
Ideally, we would verify the correctness of classifications across the entirety of the dataset, but this is not feasible from a time or cost perspective.
To indirectly assess such bias, we evaluated the NLI classification accuracy on HSplit, Wiki-BM, and Cont-BM, which were used as test sets in this study.
Given that the sets of complex sentences and simple sentences are all examples of correct, they should be classified as entailment. 
Therefore, analyzing the percentage of these instances that are classified into labels other than entailment may reveal potential biases introduced by the current NLI classifier-based filtering process.

To this end, we conducted a three-way classification of each test set using the NLI classifier (DeBERTa-xxl) and provide the results in Table~\ref{tab:nli-bias}.
From the table, it can be seen that for HSplit and Wiki-BM, all cases could be classified as Entailment.
Since both HSplit and Wiki-BM are datasets created from Wikipedia sentences, it can be inferred that the NLI classifier used in this study (DeBERTa-xxl) is capable of performing NLI classification with high accuracy on WikiSplit as well, which is also created from Wikipedia sentences.

While the classification accuracy for Cont-BM was slightly lower than that for HSplit and Wiki-BM, it was still sufficiently high. 
This difference can be attributed to Cont-BM being created from contract sentences, which belong to a domain different from Wikipedia.
This suggests that sentences that are less Wikipedia-like might have a slightly better chance of being filtered out from WikiSplit++.
However, since the classification accuracy is high enough, we believe that the impact of bias from the classifier is not significant.

\section{Conclusion}

This paper proposed a simple and practical data refinement approach for Split and Rephrase.
First, we removed unreliable training instances, i.e., pairs of complex and corresponding simple ones where the complex sentence does not entail the simple sentences, to suppress hallucinations. 
Second, we reversed the order of simple sentences in the training dataset to prevent generating complex sentences as is.
We produced WikiSplit++ by applying the data refinement for WikiSplit and then trained encoder-decoder models with it.
Manual and automatic results obtained from HSplit, Wiki-BM, and Cont-BM demonstrate that our data refinement suppresses hallucinations caused by contradictions between complex and simple sentences and increases the number of splits.
Furthermore, our data refinement has sufficient generality through experimental results on MinWikiSplit and BiSECT.

\nocite{*}
\section{Bibliographical References}\label{sec:reference}

\bibliographystyle{lrec-coling2024-natbib}
\bibliography{lrec-coling2024}

\end{document}